\documentclass[runningheads]{llncs}
\usepackage{graphicx}
\usepackage{hyperref}
\begin{document}
\title{Capturing Emerging Complexity in Lenia}
%
%
\author{Sanyam Jain\inst{1} \and
Aarati Shrestha\inst{1} \and
Stefano Nichele\inst{1,2}}
\authorrunning{S. Jain et al.}
%
\institute{Østfold University College, Halden, Norway \and
Oslo Metropolitan University, Oslo, Norway
\\
\email{\{sanyamj,aaratis,stefano.nichele\}@hiof.no}}
\maketitle              

\begin{abstract}
This work investigates the emergent complexity in Lenia, an artificial life platform that simulates ecosystems of digital creatures. Lenia's ecosystem consists of a continuous cellular automaton where simple artificial organisms can move, grow, and reproduce. Measuring long-term complex emerging behavior in Lenia is an open problem. Here we utilize evolutionary computation where Lenia kernels are used as genotypes while keeping other Lenia parameters, such as the growth function, fixed. First, we use Variation over Time as a fitness function where higher variance between the frames is rewarded. Second, we use Auto-encoder based fitness where variation of the list of reconstruction loss for the frames is rewarded. Third, we perform a combined fitness where higher variation of the pixel density of reconstructed frames is rewarded. Finally, after performing several experiments for each fitness function for 500 generations, we select interesting runs for an extended evolutionary time of 2500 generations. Results indicate that the kernel's center of mass increases with a specific set of pixels and the overall complexity measures also increase. We also utilize our evolutionary method initialized from known handcrafted kernels. Overall, this project aims at investigating the potential of Lenia as ecosystem for emergent complexity in open-ended artificial intelligence systems.

\end{abstract}


\section{Introduction}

Open-endedness is considered an important feature in Artificial Intelligence (AI) \cite{b0,b1}, because it enables the development of more flexible, creative, and autonomous systems that can solve a wider range of tasks. It also facilitates the emergence of unexpected and potentially useful behaviors that may not have been anticipated by human designers. In addition, open-endedness may allow AI systems to continually learn and adapt to changing environments. Lenia is an artificial life \cite{b2,b3} where simple rules can give rise to complex behavior in a digital system. It is a digital simulation that uses a continuous cellular automaton (CA) to generate complex patterns. Lenia differs from traditional CA in that its cells are not limited to discrete states like "on" or "off". Instead, each cell can take on a continuous range of values, which allows for more fluid and organic patterns to emerge. Lenia is also unique in that it allows for a wide range of parameters to be adjusted, such as the size and shape of the grid, the rules for how cells interact (kernel), and the speed at which the simulation runs (growth function). More in general, cellular automata (CA) have been recently shown to be well suited for AI tasks \cite{b12,b13}. In \cite{b4}, a variant of CA called Neural Cellular Automata (NCA) has been used to develop a control system for a cart-pole agent, demonstrating stable behavior over many iterations and exhibiting regeneration and robustness to disruptions. Further, extensions of classical Lenia have been proposed, such as Sensorimotor Lenia \cite{b5}, Flow Lenia \cite{b6}, and Energy based Particle Lenia \cite{b7}, which show advanced physics, chemistry and biology of the matter. In this paper, we investigate the emergent complexity in Lenia using an evolutionary approach. 


\section{Lenia}

\begin{figure}[t]
    \centering
    \includegraphics[width=0.9\textwidth]{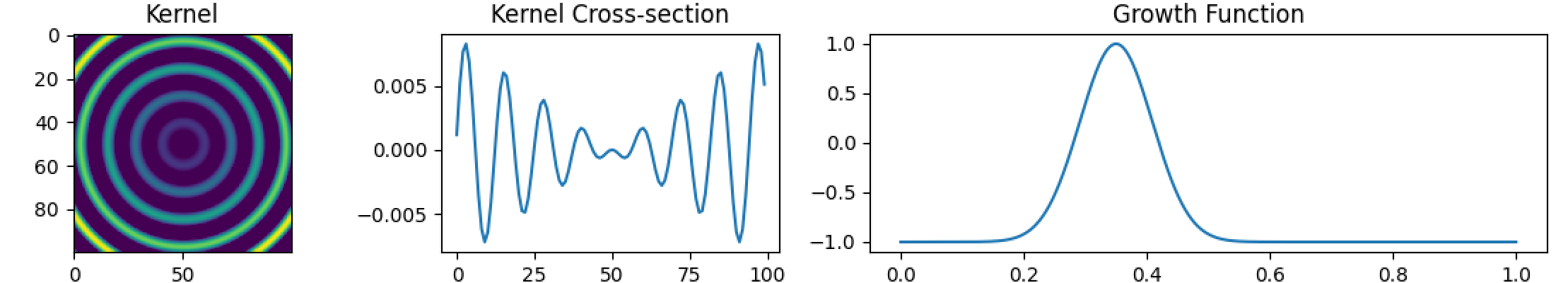}
    \caption{Visualising Kernel, Gaussian Kernel Cross-Section Function and Gaussian Growth Function}
    \label{fig:kg}
\end{figure}

Lenia \cite{b2,b3} is a digital life simulation software that utilizes a unique approach to simulate various patterns and behaviors through the use of an abstract concept called the "growth function." The growth function in Lenia represents the process of growth and decay of various shapes and structures, which is governed by two important parameters: $\mu$ and $\sigma$.

The $\mu$ parameter in Lenia's growth function is the mean of the function, which determines the overall behavior of the shapes in the simulation. The $\sigma$ parameter, on the other hand, represents the standard deviation of the function. It controls the degree of randomness in the growth and decay process. Together, the $\mu$ and $\sigma$ parameters of the growth function enable Lenia to generate a wide range of complex patterns. Over 500 species have been discovered so far \cite{b11}, with each residing in a unique combination of these parameters. 

Similarly, the Gaussian kernel is used for smoothing and blurring the simulation board, and it is a circular matrix that is centered around the current cell. The number of smooth rings or "shells" in the Gaussian kernel is determined by the "peaks" parameter, which specifies the amplitude of the peaks for each shell. A detailed view of kernel and growth function can be seen in Figure \ref{fig:kg}. For more details on Lenia, please refer to \cite{b2,b3}.

\section{Methodology}

We propose modified versions of two known approaches called compression based metric and deviation based approach to evaluate cellular automata (CA) \cite{b10}. Previously these approaches have been studied on classical CA, we propose novel modified approaches namely, AutoEncoder (AE) based compression metric, Variation over Time (VoT) based deviation metric and a completely novel technique called as Auto Encoder Variation Over Time (AEVoT) which is built on top of known AE and VoT approaches. We describe each of them as follows:

\subsection{Variation Over Time (VoT) fitness}
\begin{figure}[ht]
    \centering
    \includegraphics[width=0.9\textwidth]{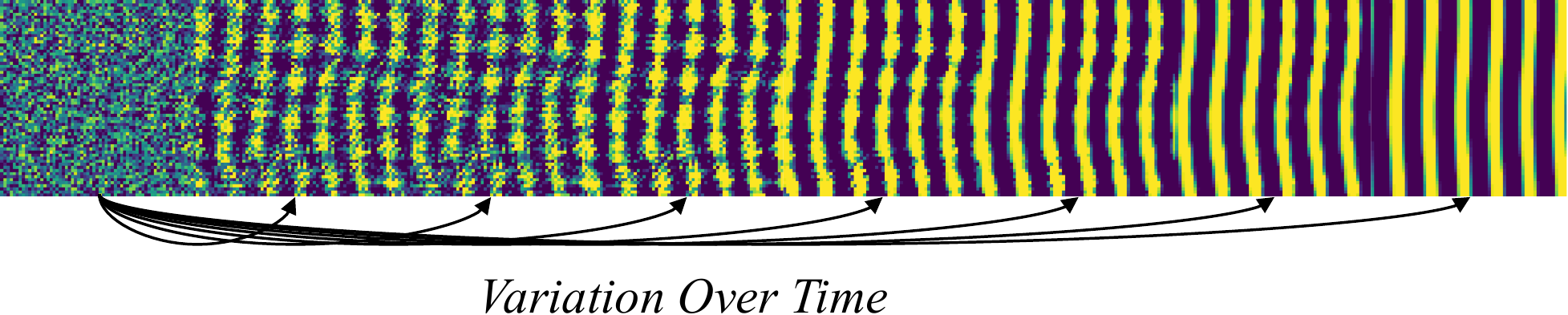}
    \caption{Variation over time approach to measure complexity}
    \label{fig:vot}
\end{figure}

One way to quantify complexity in Lenia is to measure the variation in the state of the board over time. One approach is to measure the number of active cells on the board at each time step. The number of active cells can provide a rough estimate of the complexity of the system, as more active cells typically indicate a higher degree of complexity. However, it is important to note that the number of active cells can vary significantly between different time steps, and may not be a reliable measure of complexity in all cases. 
A threshold (a hyper-parameter) is used to determine the alive state of the grid. For pixel values smaller than threshold, cells are considered dead. Count from each grid timestep is used to calculate the standard deviation. Deviation is a measure of the randomness or differences in states of a system, and can be used to quantify the variation of complexity in a system over time. The VoT approach is particularly useful in identifying temporal patterns and trends in the system's behavior, which can be indicative of its underlying complexity. An overview is shown in Figure \ref{fig:vot}

\subsection{Autoencoder (AE) based fitness}
\begin{figure}[ht]
    \centering
    \includegraphics[width=0.9\textwidth]{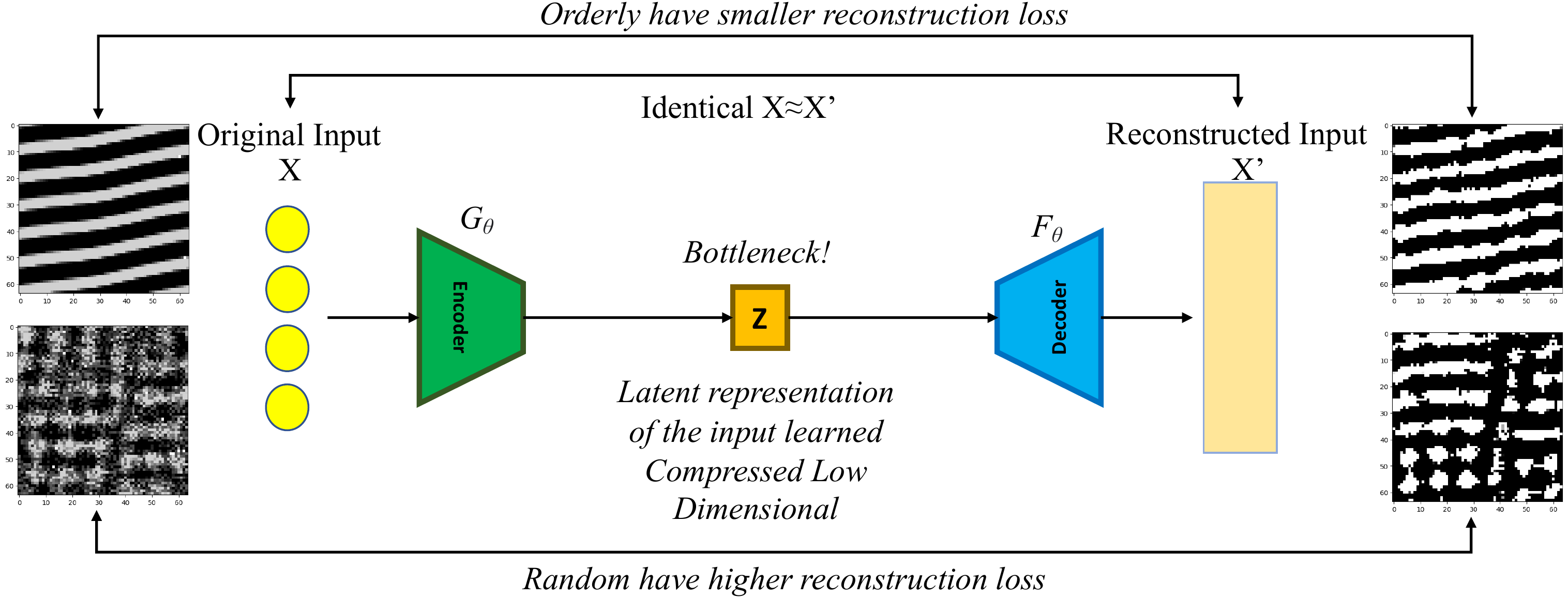}
    \caption{Auto-encoder approach to measure complexity}
    \label{fig:ae}
\end{figure}

Another approach, which uses an encoder-decoder network to compress the input images, has been proposed in \cite{b10} as a way to measure the complexity of CA. In this approach, the encoder network is trained to encode the current state of the Lenia board into a lower-dimensional feature space, while the decoder network is trained to decode the features back to the original state. The key idea is that the amount of information that is lost during the encoding and decoding process (reconstruciton loss) can be used as a measure of the complexity of the Lenia board. Our hypothesis is that if the input is complex or chaotic, it is more challenging for the decoder to reconstruct the input (high reconstruction loss), while in the ordered behaviour the reconstruction loss would be small. The input frames are reconstructed using the trained AE and the Mean Squared Error (MSE) loss is stored. The MSE is calculated between original frames and reconstructed frames. After that, the standard deviation is calculated between those errors. If the list of errors has a higher deviation, then there should be a larger variance between the frames generated using the evolved kernel. To train the encoder-decoder network, a dataset of Lenia board states is first generated. The encoder network is then trained to map each board state to a set of features that capture the key patterns and structures in the board. The decoder network is trained to reconstruct the board state from the features, with the goal of minimizing the reconstruction error. An overview is shown in Figure \ref{fig:ae}

\subsection{Auto Encoder Variation Over Time}
The Auto-Encoder based Variation over Time (AEVoT) approach combines the two previously described methods, AE and VoT, to measure the complexity of the Lenia patterns over time on reconstructed frames. It is very similar to the AE and VoT approaches, with a difference that VoT is calculated on reconstructed images from the AE. To use an AE for time series data, we can treat each time step as a separate input and train the AE to reconstruct the original sequence of data. The reconstruction loss, which is the difference between the input and the reconstructed output, can then be used to measure the complexity of the time series data. By analyzing the variation in the reconstruction loss over time, we can gain insights into the complexity of the time series. For example, if the reconstruction loss is relatively constant over time, it may indicate that the time series is relatively simple or predictable. However, if the reconstruction loss varies widely over time, it may indicate that the time series is more complex and difficult to predict.


\section{Experimental Setup and Results}



\begin{figure}[ht]
    \centering
    \includegraphics[width=0.7\textwidth]{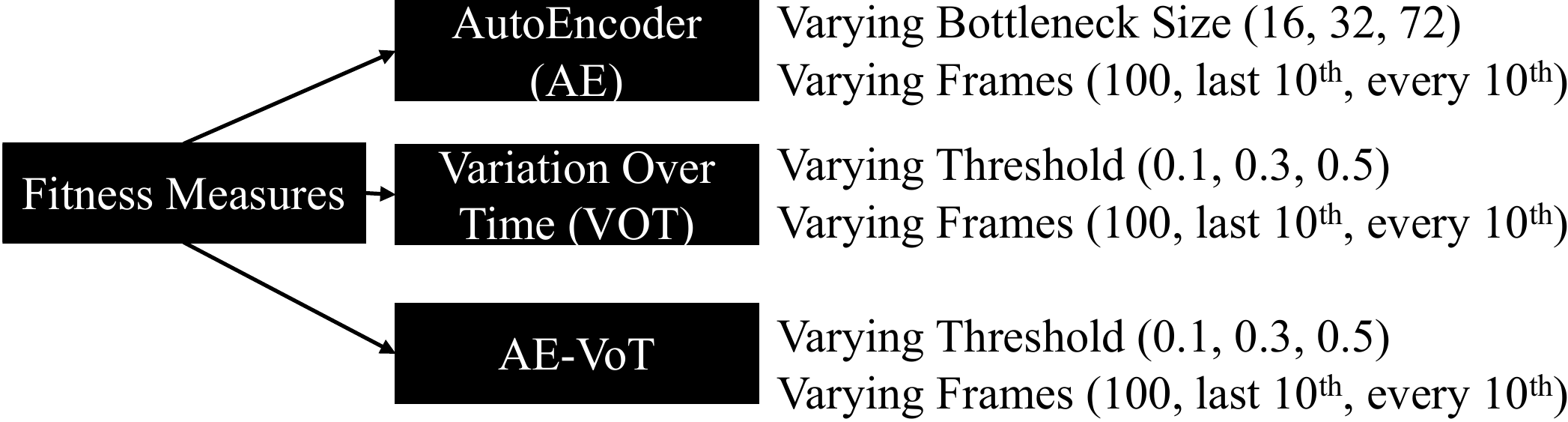}
    \caption{Overview of experiments performed}
    \label{fig:arrow}
\end{figure}

A high level overview of experimental setup is provided in Figure \ref{fig:arrow}.

We developed an evolutionary algorithm where Lenia kernels are evolved with the three fitness functions outlined above. The initial population comprises randomly initialized kernels that are applied from a randomly initialized Lenia board.
Additionally, the Lenia frames are processed in three different ways, namely "all produced frames for $t$ time steps", "every 10th frame" and "last 10 frames". The idea is to capture long term emerging complexity so the behavior should be present not only at initial frames but in later frames as well.

The selection function used in the code is a Roulette Wheel Selection method, which selects individuals based on their fitness proportionate to the sum of the fitness of all individuals in the population. 
The number of elites in the population is set to one. 


The mutation operator mutates an individual's genes by randomly altering each gene with a probability equal to the $mutation\_rate$ value. The $mutation\_rate$ value is set to 0.02, which means that each gene has a $2\%$ chance of being mutated. If the random value generated for a gene is less than $mutation\_rate$, the gene is randomly reassigned a value between 0 and 1, rounded to 3 decimal places. 

For each of the experiments, a system configuration of 64 GB of RAM, 250 GB Hard Disk space, 8vCPU, in a AWS Sagemaker ML.R5.2xLarge instance is used. It takes almost ~24 hours to finish one run of 500 generations with our configuration. For the AE training dataset, the Lenia parameters are given in Table \ref{tab:dataset}. 
The configuration for AE is provided in Table \ref{tab:ae} and the overall configuration for the Evolutionary Algorithm is provided in Table \ref{tab:ga}.

\begin{table}[htbp]
  \centering
  \caption{Dataset Specifications}
    \begin{tabular}{|l|l|}
    \hline
    \textbf{Parameter} & \textbf{Value} \\
    \hline
    Dataset Size & 3000 frames \\
    \hline
    DPI   & 50 \\
    \hline
    $Mu_G$   & 0.31  \\
    \hline
    $Sigma_G$ & 0.057 \\
    \hline
    Dt    & 0.1 \\
    \hline
    FPS   & 30 \\
    \hline
    Kernel Size & 16 \\
    \hline
    Board Size  & 64 \\
    \hline
    \end{tabular}%
  \label{tab:dataset}%
\end{table}%




\begin{table}[htbp]
  \begin{minipage}[b]{0.45\linewidth}
    \centering
    \caption{Training Configuration for AE}
    \begin{tabular}{|l|l|}
    \hline
    \textbf{Parameter} & \textbf{Value} \\
    \hline
    Image Size   & 64x64 \\
    \hline
    AE test size   & 0.30  \\
    \hline
    AE Input Size  & 64x64 \\
    \hline
    AE Hidden Size     & 36 \\
    \hline
    AE Ouptut Size    & 64x64 \\
    \hline
    Epochs & 300 \\
    \hline
    Batch Size  & 128 \\
    \hline
    \end{tabular}%
  \label{tab:ae}%
  \end{minipage}\hfill
  \begin{minipage}[b]{0.45\linewidth}
    \centering
    \caption{Configuration for Genetic Algorithm}
    \begin{tabular}{|l|l|}
    \hline
    \textbf{Parameter} & \textbf{Value} \\
    \hline
    Kernel size  & 16 \\
    \hline
    Board size   & 64 \\
    \hline
    Mutation rate   & 0.02 \\
    \hline
    Population size     & 10 \\
    \hline
    Generation  & 500 \\
    \hline
    Elites & 1 \\
    \hline
    \end{tabular}%
  \label{tab:ga}%
  \end{minipage}
\end{table}%

\begin{figure*}[htbp]
    \centering
    \includegraphics[width=1\textwidth]{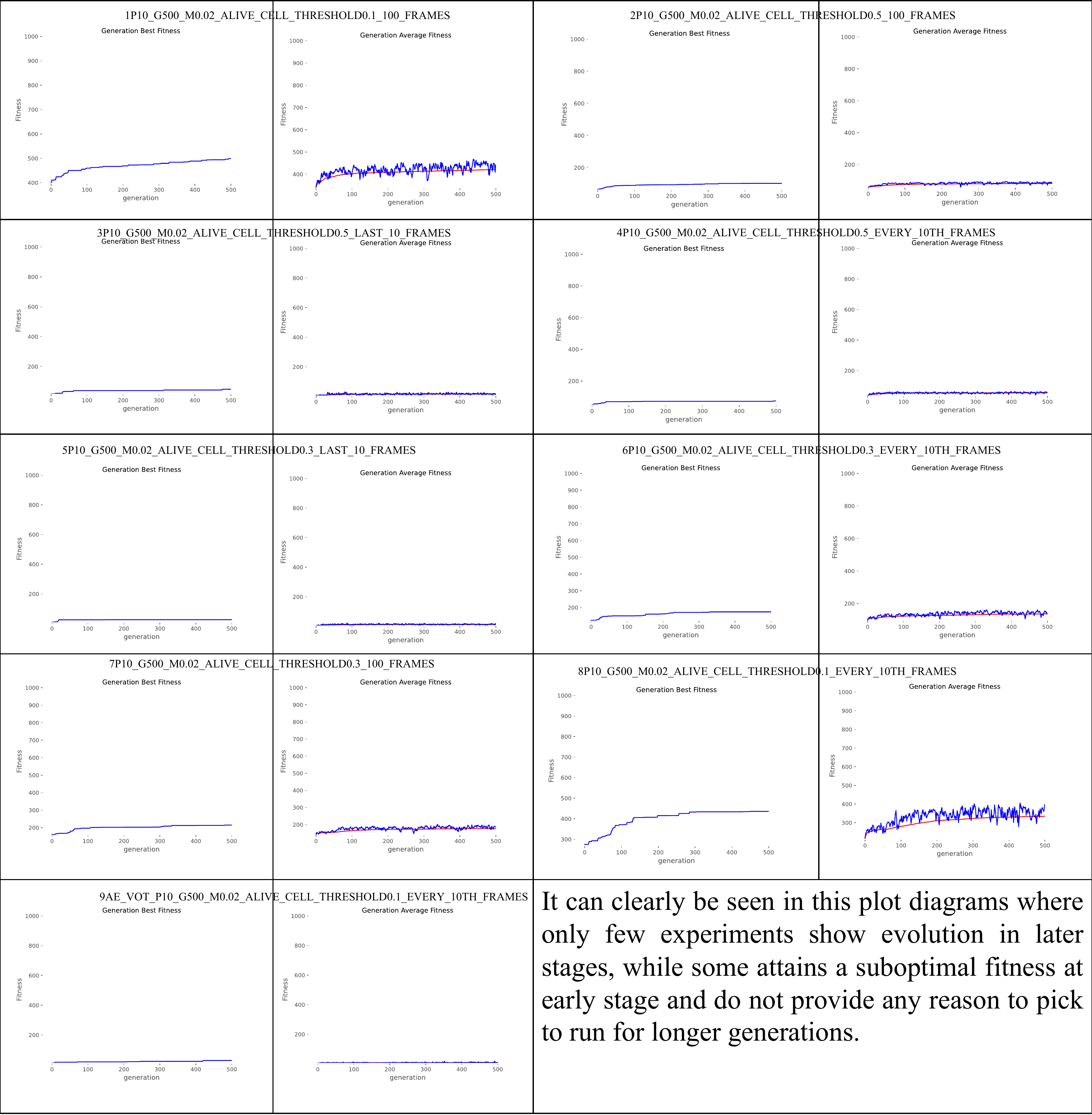}
    \caption{Plot data for Variation over Time fitness. Configuration should be read as for example 1AE\_36bottleneck\_allframe Autoencoder fitness for bottleneck size of 36 ran for all 100 frames. Y-axis has upper limit of 1000.}
    \label{fig:1t}
\end{figure*}

\begin{figure*}[htbp]
    \centering
    \includegraphics[width=1\textwidth]{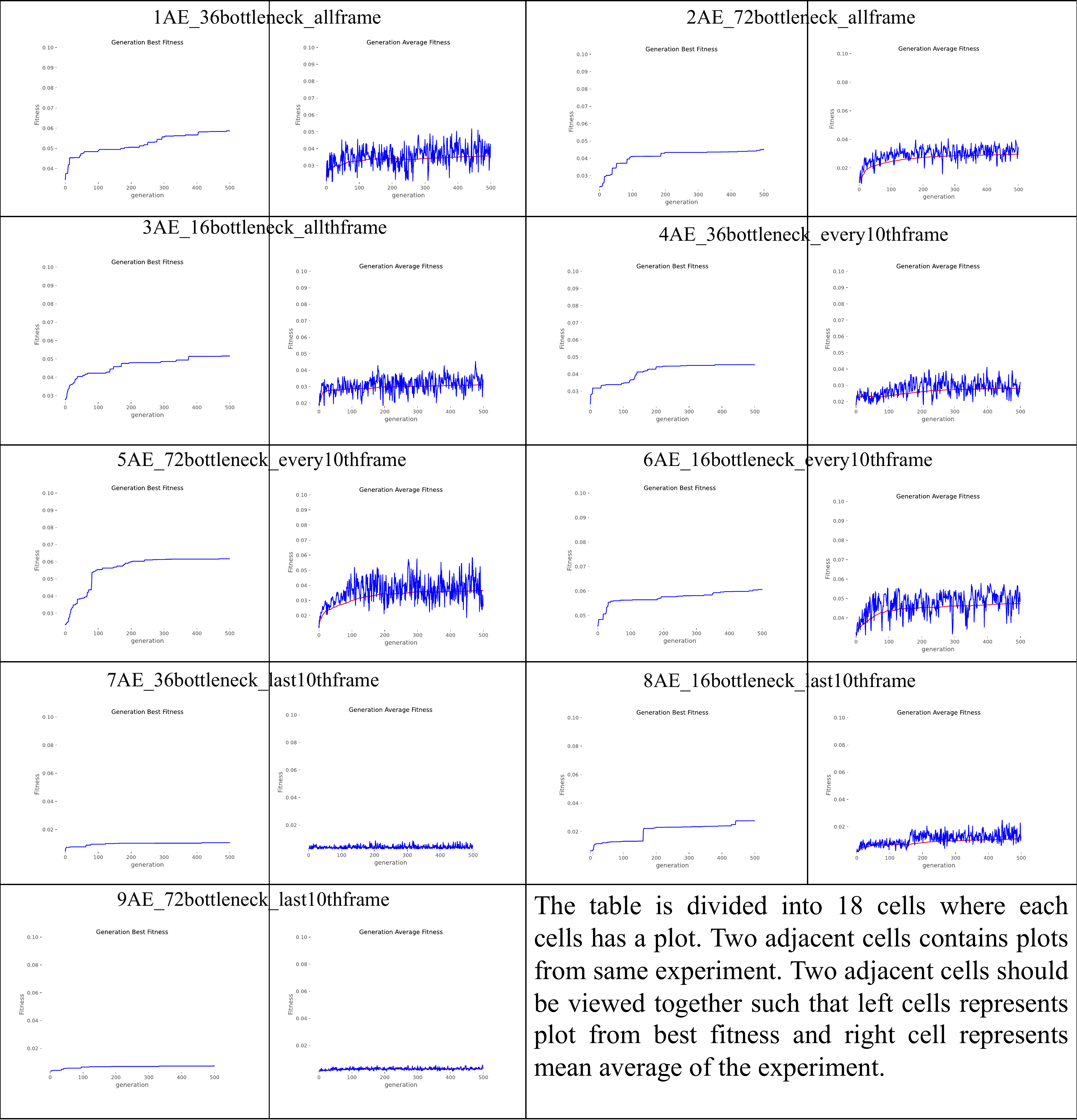}
    \caption{Plot data for AutoEncoder fitness. Configuration should be read as for example 1P10\_G500\_M0.02\_ALIVE\_CELL\_THRESHOLD 0.1\_100\_FRAMES should be read as population size (P) as 10 ran for 500 generations (G), with mutation rate (M) of 0.02, active cell threshold 0.1 for 100 frames. Y-axis has upper limit of 0.1.}
    \label{fig:2t}
\end{figure*}

\begin{figure*}[htbp]
    \centering
    \includegraphics[width=1\textwidth]{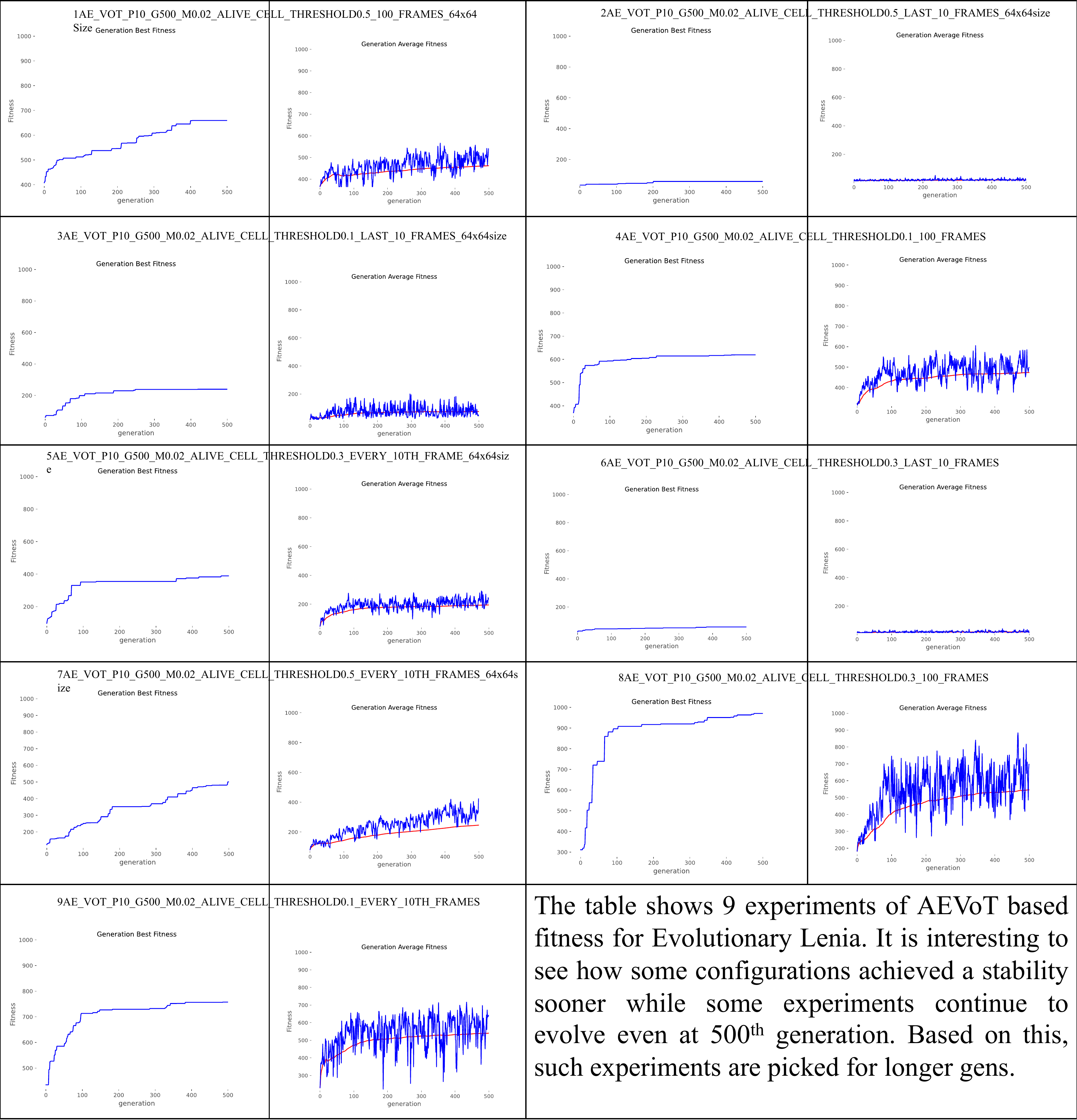}
    \caption{Plot data for AutoEncoder with Variation over Time fitness. Configuration should be read as for example 1AE\_VOT\_P10\_G500\_M0.02\_ALIVE\_CELL\_THRESHOLD0.5\_100\_FRAMES should be read as population size (P) as 10 ran for 500 generations (G), with mutation rate (M) of 0.02, active cell threshold 0.5 for 100 frames with 64 as input image size. Y-axis has upper limit of 1000.}
    \label{fig:3t}
\end{figure*}

All experimental results for the different combinations of parameters are displayed for 500 generations. In addition best fitness and average fitness are also plotted in the Figure \ref{fig:1t} for VoT, Figure \ref{fig:2t} for AE and Figure \ref{fig:3t} for AEVoT.


Once all the experiments in Figures \ref{fig:1t}, \ref{fig:2t} and \ref{fig:3t} were finished, we analysed and selected one experiment from AEVoT and one from VoT on the basis of their promising performance. We also found that AE itself is not able to perform well as per the experiments shown in Figure \ref{fig:2t}. AE alone could not capture the complex of Lenia behaviours because it merely reconstructs the frames and tries to measure MSE loss over the reconstruction. AEVoT achieved the best fitness performance when ran for 2500 generations. We found, for AEVoT, Active cell threshold 0.5 and every $10^{th}$ frame is working better than other configurations (with fitness value of 500.27). Moreover in VOT, Active cell threshold 0.1 for every $10^{th}$ frame is working particularly well (with fitness value of 435.91). 

\begin{figure}[ht]
    \centering
    \includegraphics[width=0.6\textwidth]{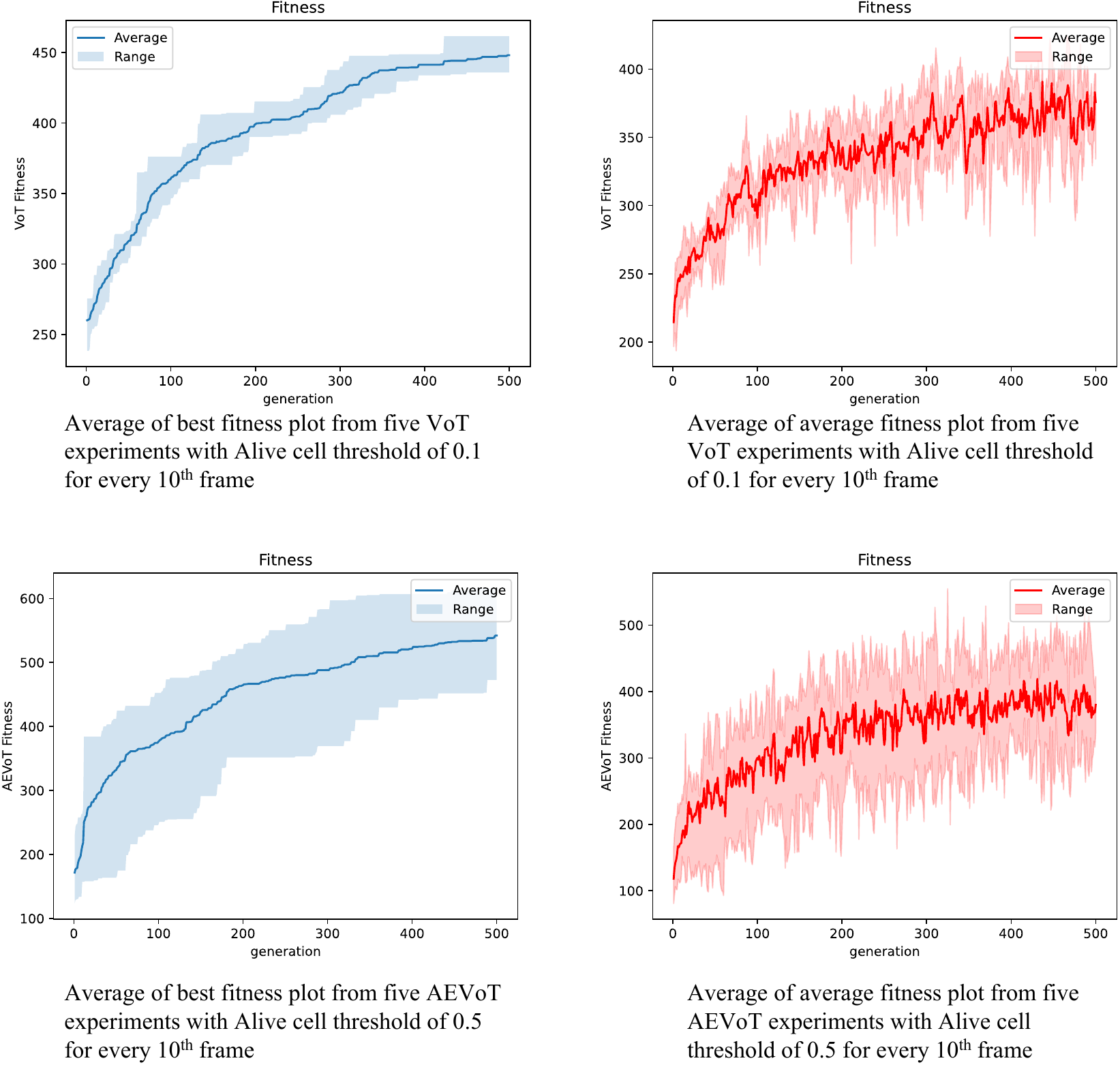}
    \caption{Five-fold-averaging of best and average fitness values of five experiments with same configurations shown in Figure \ref{fig:2500}. First row shows VoT run over 500 generations (5 times) using Alive threshold of 0.1 and measuring over every tenth frame. Second row shows AEVoT run over 500 generations (5 times) using Alive threshold of 0.5 and measuring over every tenth frame.}
    \label{fig:fivefold}
\end{figure}

\begin{figure}[ht]
    \centering
    \includegraphics[width=0.6\textwidth]{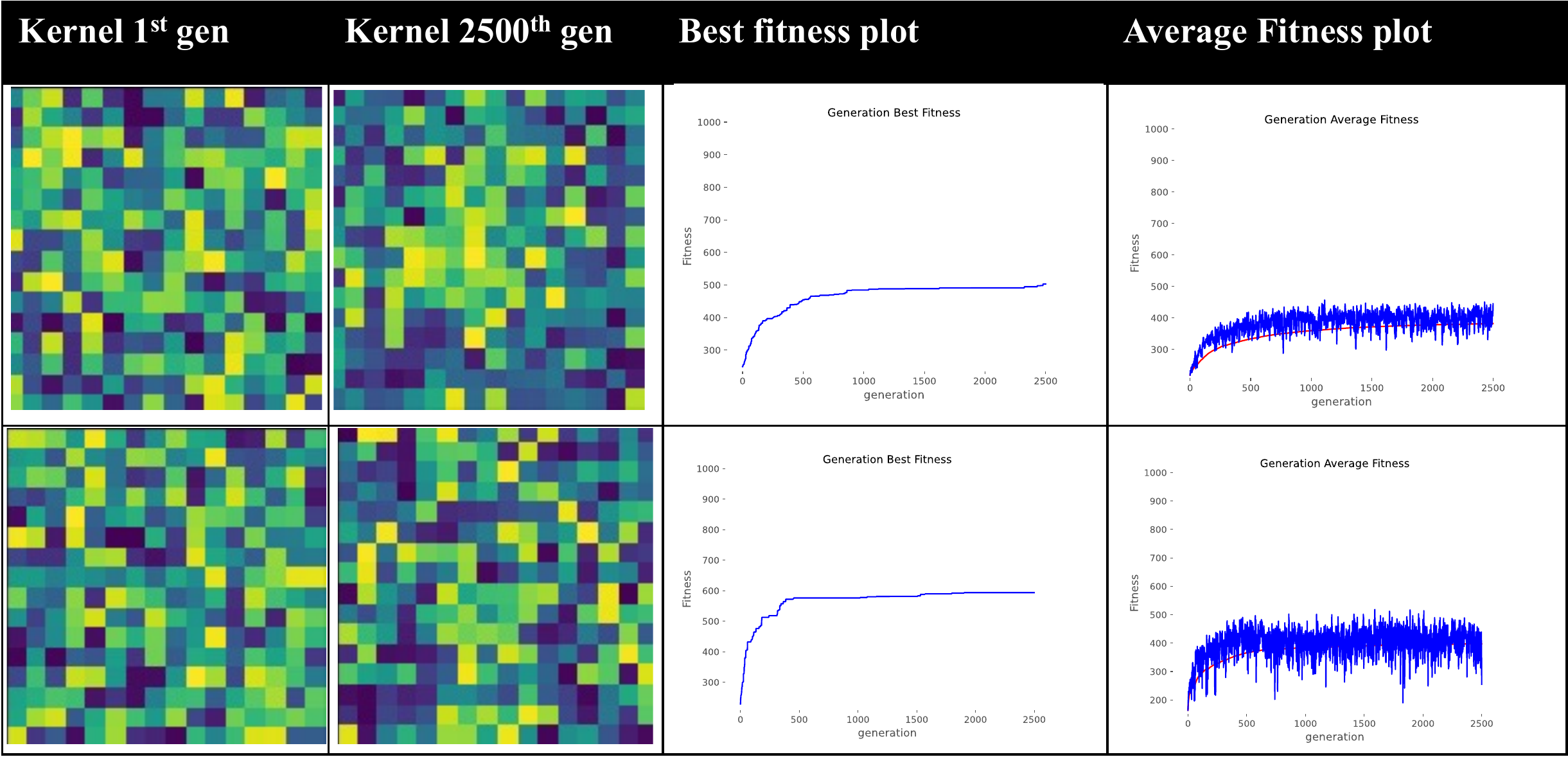}
    \caption{Best and average fitness plots with kernel visualisations for the selected configuration running for 2500 generations. First row is VoT based simulation with configuration picked from Figure \ref{fig:1t} $8^{th}$ row and second row is AEVoT based simulation with configuration picked from Figure \ref{fig:3t} $7^{th}$ row }
    \label{fig:2500}
\end{figure}

It may be observed that for variation over time, active cell threshold should be small and has to be calculated for a time delta of a significant value over span of total time steps. In our case, we have a total time step of 100 where time delta is 10 and hence fitness is being calculated for every $10^{th}$ frame. Hence, we provide a detailed analysis for these two experiments with kernel visuals from generation 1 to generation 2500 along with their long term best and average fitness plots in Figure \ref{fig:2500}. It is important to note that after 2500 generations, the kernel centre of mass should have large pixel density of Lenia behaviour promoting pixel groups (can be seen in "yellowish" color map in the kernel plot, however all the dark pixels with dark blue color are outlined towards boundary in the plot). Finally we plot five-fold-average for best and average fitness values of the same experiments shown in Figure \ref{fig:2500} ran on five different run-times, which is shown in Figure \ref{fig:fivefold}

\begin{figure}[ht]
    \centering
    \includegraphics[width=0.9\textwidth]{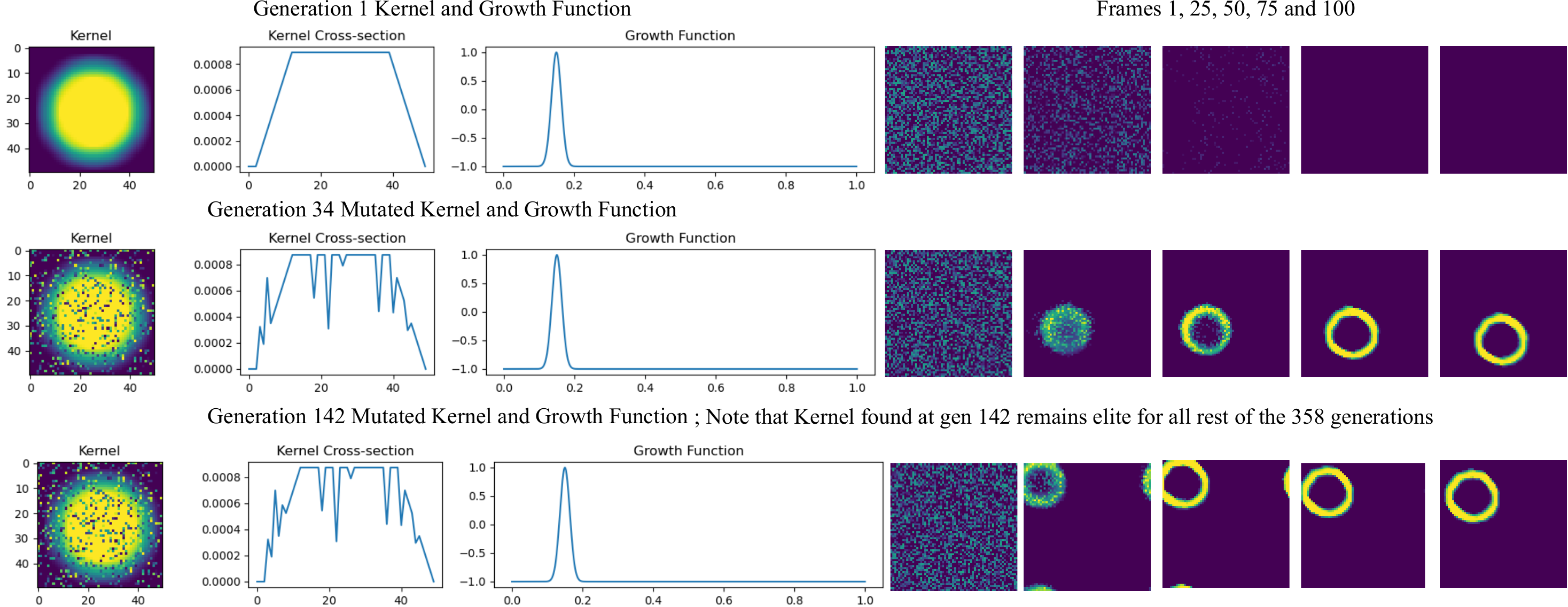}
    \caption{Complex behaviour emerging in Evolutionary Lenia with the known kernel and its mutation for 500 generations}
    \label{fig:spherical}
\end{figure}

\begin{figure}[ht]
    \centering
    \includegraphics[width=0.7\textwidth]{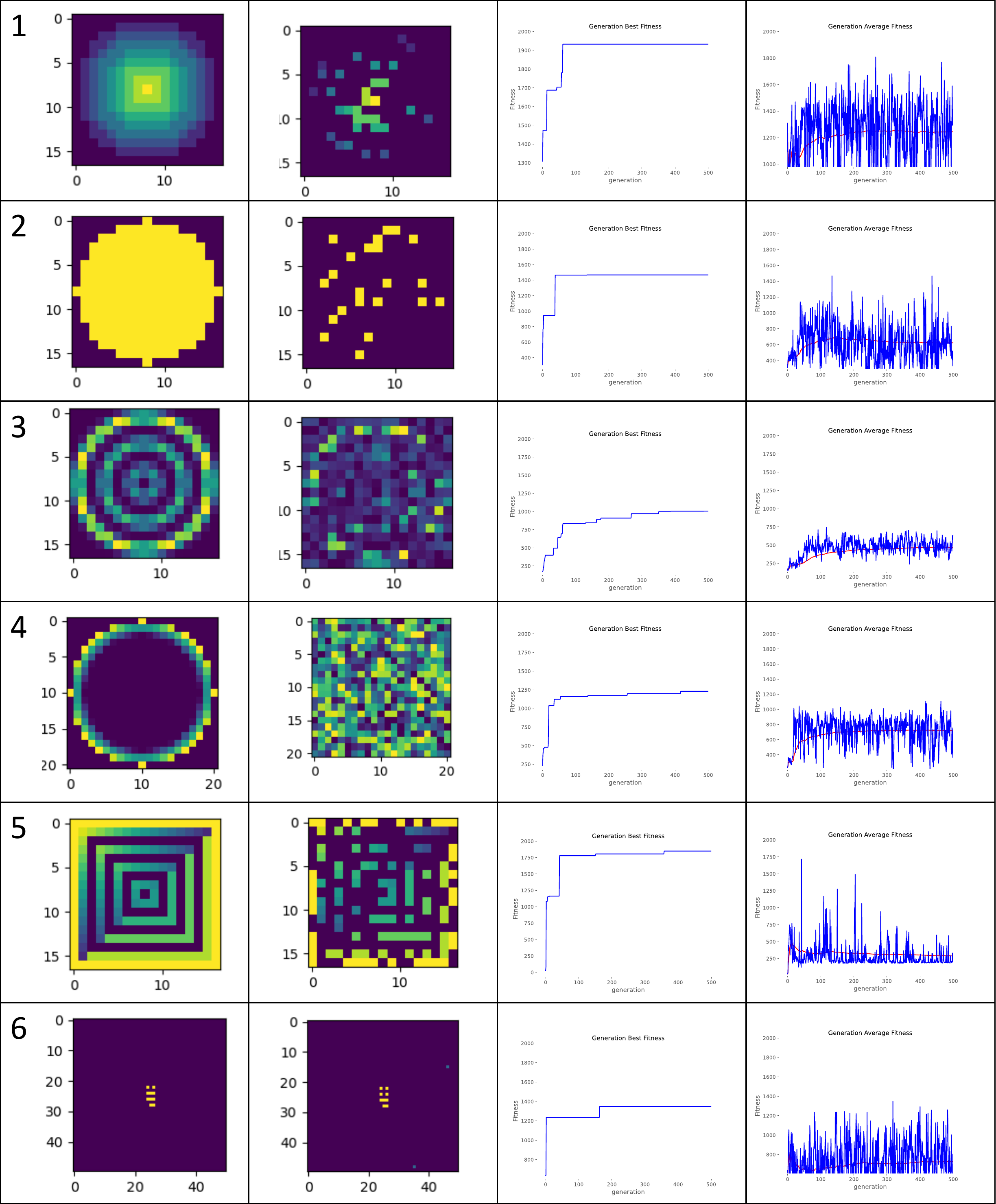}
    \caption{Running 500 generation based simulation for known kernels. The configuration is kept same as the simulation shown in Figure \ref{fig:2500} second row. First column shows the initial known kernel, second column shows mutated kernel after 500 generations. Additionally, third column shows best fitness plot and fourth column shows average fitness plot with cumulative mean-average line in between. (Plots have a y-limit of 2000)}
    \label{fig:known}
\end{figure}

To compare our experiments with handcrafted kernels or known kernel, we perform six experiments keeping growth function and the rest of the configuration the same. Known kernels are very robust to perturbations. Even after doing mutations for these kernels they show similar behaviour and complexity measurement remains high, which can be shown by the generation fitness curve depicted in Figure \ref{fig:known}. At the first place, even after mutations, there are very less changes happened in the fitness measurement value. In other words, we let evolution run for 500 generations and each generation has mutation impact of 2 pixels. However, the growth of fitness happens slowly in those generations, which allows the behavior to remain robust even after perturbations. For example, with the kernel configuration and best plots in the $3^{rd}$ column shown in Figure \ref{fig:known}, the growth has happened only at first few generations and then it became stable very soon. Finally, after handcrafting a kernel with a smoothed spherical Gaussian pixel distribution which is shown in Figure \ref{fig:spherical}, along with the popular known growth function, shows particularly complex and self-organising behaviour. Metaphorically, it shows a ring like bacteria pattern emerging from a random grid. All the results from such a kernel are shown in Figure \ref{fig:spherical}. The configuration used here is the same used in Figure \ref{fig:known}.

Full code with results is openly available at this link: \url{https://s4nyam.github.io/evolenia}.  All experimental visualisations and animations were programmatically processed to produce an overall summary shown on our YouTube channel: \url{www.bit.ly/leniaonyt}. The GitHub repository is \url{www.github.com/s4nyam/APCSP}.

\section{Conclusions}

In this work, we evolve Lenia kernels to identify complex emerging behaviour for a specific set of parameters in a large and wide existing parameter-space. To explore such wide space of parameters, we performed multiple evolutionary experiments that exploited mutations of kernel pixels while starting from a randomly initialised kernel, and keeping growth function fixed. We measured complexity using three techniques, namely AE, VoT and AEVOT. Once we found a specific set of configuration performing well, we ran for longer generations and achieved a rather different behaviour emerging in Lenia when compared with the initial random behaviour. Lenia, as a computational system, provides a platform for exploring emergent behaviors that mimics some of the properties of living systems, such as self-organization, pattern formation, and adaptation to changing environments. By studying these behaviors in Lenia, we hope to gain insights into key underlying mechanisms that may be beneficial for the development of a more general AI. 


\end{document}